\documentclass[12pt]{article}

\usepackage{arxiv}

\usepackage[utf8]{inputenc} 
\usepackage[T1]{fontenc}    
\usepackage{hyperref}       
\usepackage{url}            
\usepackage{booktabs}       
\usepackage{rotating}
\usepackage{tablefootnote}

\usepackage{multirow}
\usepackage{ragged2e}

\usepackage{amsfonts}       
\usepackage{nicefrac}       
\usepackage{microtype}      
\usepackage{lipsum}
\usepackage{graphicx}
\usepackage{indentfirst}
\usepackage{array}
\usepackage{gensymb}
\graphicspath{ {./images/} }

\usepackage{fancyhdr}
\fancyheadoffset{0pt}
\usepackage{hyperref}
\hypersetup{
    colorlinks=true,
    linkcolor=blue,
    filecolor=magenta, 
    citecolor=blue,     
    urlcolor=cyan,
}

\def\keywordname{{\bfseries Keywords:}}%
\def\keywords#1{\par\addvspace\medskipamount{\rightskip=0pt plus1cm
\def\and{\ifhmode\unskip\nobreak\fi\ $\cdot$
}\noindent\keywordname\enspace\ignorespaces#1\par}}

\providecommand{\keywords}[1]
{
  \small	
  \textbf{\textit{Keywords---}} #1
}

\title{{\sc {\sc PemNet}}: A Transfer Learning-based Modeling Approach of High-Temperature Polymer Electrolyte Membrane Electrochemical Systems}

\author
{L. A. Briceno-Mena$^{1}$, C. G. Arges$^{1}$, J. A. Romagnoli$^{1\ast}$\\
\\
\normalsize{$^{1}$Cain Department of Chemical Engineering, Louisiana State University,}\\
\normalsize{Baton Rouge, LA 70803, USA}\\
\\
\normalsize{$^\ast$To whom correspondence should be addressed; E-mail:  jose@lsu.edu.}
}

\date{}

\begin{document}
\maketitle
\begin{abstract}
Widespread adoption of high-temperature polymer electrolyte membrane fuel cells (HT-PEMFCs) and HT-PEM electrochemical hydrogen pumps (HT-PEM ECHPs) requires models and computational tools that provide accurate scale-up and optimization. Knowledge-based modeling has limitations as it is time consuming and requires information about the system that is not always available (e.g., material properties and interfacial behavior between different materials). Data-driven modeling on the other hand, is easier to implement, but often necessitates large datasets that could be difficult to obtain. In this contribution, knowledge-based modeling and data-driven modeling are uniquely combined by implementing a Few-Shot Learning (FSL) approach. A knowledge-based model originally developed for a HT-PEMFC was used to generate simulated data (887,735 points) and used to pretrain a neural network source model. Furthermore, the source model developed for HT-PEMFCs was successfully applied to HT-PEM ECHPs - a different electrochemical system that utilizes similar materials to the fuel cell. Experimental datasets from both HT-PEMFCs and HT-PEM ECHPs with different materials and operating conditions (\char`~~50 points each) were used to train 8 target models via FSL. Models for the unseen data reached high accuracies in all cases (rRMSE between 1.04 and 3.73\% for HT-PEMCs and between 6.38 and 8.46\% for HT-PEM ECHPs). 
\end{abstract}


\section{Introduction}
Electrochemical systems for energy conversion and storage, as well as chemical manufacturing, powered on solar and wind are central to reducing greenhouse gas emissions in addition to attaining a more sustainable way of living. For example, fuel cell electric vehicles using green hydrogen fuel pose a great opportunity to curtail greenhouse gas emissions that hail from transporting goods via heavy duty vehicles (e.g., trucking), marine ships, aviation, and trains \cite{RN1, RN2, RN3}. Green hydrogen is derived from water electrolysis that use electrical energy from renewable energy sources (e.g., solar and wind) and water as a reactant. Hydrogen is a vital feedstock to the production of fertilizers (i.e., ammonia) via Haber-Bosch process in addition to refining metals (e.g., steel) and carrying out numerous chemical operations such as hydrogenation reactions. Hydrogen is also important for use as coolant for thermal-electric power plants due its high specific heat capacity. For applications outside water electrolysis and fuel cell electric vehicles, electrochemical hydrogen pumps (ECHPs) are central to purifying hydrogen, compressing hydrogen and distributing hydrogen for chemical manufacturing. Greater adoption of FCEVs and ECHPs platforms necessitate further cost reductions and improvements in efficiency and stability. 

In recent years, high-temperature polymer electrolyte membrane (HT-PEM) fuel cells and ECHPs have experienced a renewed interest since the advent of ion-pair PEM architectures that expand the temperature range for these platforms and tolerance to higher levels of humidity\cite{RN4, RN5, RN6, RN7}. Pairing HT-PEMs with phosphonated ionomer electrode binders in the past year have resulted in high performing fuel cells and ECHPs. For example, a peak power of 1.7 W cm\textsuperscript{-2} has been achieved for HT-PEMFCs using hydrogen and oxygen\cite{RN5, RN6, RN7, RN8}, while a HT-PEM ECHP has shown 1 A cm\textsuperscript{-2} at 55 mV\cite{RN7}. A central advantage of HT-PEM fuel cells and hydrogen pumps is their ability to tolerate carbon monoxide (CO) in the hydrogen stream as CO adsorption at temperatures of 200 °C or above is diminished\cite{RN9}. Over 90\% of all hydrogen is derived from steam methane reforming (SMR) that leads to mixtures of hydrogen and CO. Electrochemical processes that can tolerate CO allow the use of low-cost hydrogen because hydrogen from SMR is about 2x to 3x cheaper than hydrogen from water electrolysis. Furthermore, an added benefit of higher temperature operation for fuel cell electric vehicle stacks is that it simplifies heat management as the larger temperature gradient favors greater heat rejection and the ion-pair HT-PEMs do not require humidification for ionic conduction. This latter attribute eliminates an external humidifier, an ancillary unit that adds cost, and thus the removal ancillary units simplify the balance of plant for the fuel cell system. Despite these extraordinary developments in a relatively short-period of time, further maturation of HT-PEMFC and HT-ECHP platforms is needed to reduce platinum group metal (PGM) loading to reduce overall system capitals costs while maintaining or exceeding the demonstrated performance.  

Lowering PGM loadings requires greater electrocatalyst utilization in the electrode layers of HT-PEMFCs and HT-PEM ECHPs while also curtaining interfacial resistances in the electrode layers related to reactant, electron, and ion transport as well as ion transport to the HT-PEM. New materials, such as ionomer electrode binders, offer tremendous opportunity to enhance electrocatalyst utilization while co-currently addressing interfacial resistances; however, the material properties and how they affect interfacial reaction kinetics and transport behavior is poorly understood under at various cell operating conditions (i.e., temperature, pressure, stoichiometric ratio, etc.). This poor understanding typically leads to an ‘Edisonian’ approach of testing new materials and observing performance with little attention given to modeling the systems and trying to bridge the gap between materials properties and device level performance. Additionally, the approach of synthesize and test gives limited knowledge in short period of time and stymies the development and optimization of system performance to specified constraints (e.g., PGM loadings). Accelerating the maturation of HT-PEM fuel cells and hydrogen pumps with phosphonated ionomer electrode binders and ion-pair membranes requires a comprehensive and computationally inexpensive model usable for optimization but also capable of capturing the properties of new materials and their influence on cell performance.

Since a physical understanding of the system is required for this approach, the modeling task is complex and time consuming, and sometimes not all necessary information is available for capturing all the relevant interactions among the variables being modeled\cite{RN11}. Furthermore, for knowledge-based models, new experimental findings relevant to a specific component of the system cannot be easily incorporated into an existing model, and data on the impact of these on the overall system must be collected as well. Conversely, the data-driven approaches, which includes Machine Learning (ML), are used to obtain a model from the raw experimental data\cite{RN12}. This approach has the advantage that given enough data, the model could potentially represent all the interactions among the system’s variables within the range of the data. However, the amount and variety of data required for the actual implementation of this approach in nascent HT-PEM electrochemical devices can be cost-prohibitive. This latter issue, also called scarce availability of data is prevalent in many engineering applications and prevents the full exploitation of data-driven modeling. Recently, we reported a framework that exploits the benefits of both approaches by hierarchically combining data-driven modeling for materials behavior and knowledge-based modeling for fuel cell device performance\cite{RN13}. While this approach helped reduce the time and the amount of experimental data required for the model development, some aspects like the incorporation of new knowledge and the flexibility of the modeling approach needs further improvement. In this work, we demonstrate that Transfer Learning can be used to address these shortcomings.

Transfer Learning (TL)\cite{RN14} is a ML technique derived from the notion of transfer of learning first proposed in the educational research context\cite{RN15}. In TL, a data-driven model previously trained (general training) for a given task (source domain) is used as the base to build a model for a new task (target domain), with less data being required for the new training stage (task-specific training). A useful extension of this technique is the so-called few-shot learning (FSL) in which the task-specific training stage uses a very small amount of data (i.e., on the order of \(1\times 10^1\))\cite{RN16}. By using FSL, new findings can be easily incorporated with little experimental cost. Furthermore, the information contained in a model for a given device can be readily transferred to a model for a different, -yet similar device architecture\cite{RN17, RN18}.

Although the task-specific training can be carried out with a small data set from the target domain, the general training still requires large amounts of data from the source domain. In some applications like image or speech recognition, an existing large dataset can be used to pretrain the model and then proceed with the transfer\cite{RN19, RN20}. For the case of PEM-based electrochemical systems however, such data is not available and must be generated. Using an existing knowledge-based model, large and balanced datasets can be produced to support the pretraining stage. Furthermore, since the datasets come from a knowledge-based simulation theoretical knowledge can be incorporated into the new model development.

In this contribution, we present PEMNET, a TL approach for the modeling of HT-PEMFCs and HT-PEME CHPs that uses a knowledge-based model validated for our previously reported HT-PEMFC data as source domain. First, a strategy to leverage a knowledge-based model to generate a useful dataset is discussed. Then, the applicability of PEMNET to model both fuel cell and hydrogen pump devices is demonstrated by obtaining models for 6 HT-PEMFCs with different membrane and ionomer binder chemistries, and for two HT-PEM ECHP with different ionomer binders.

\section{Methods}

\begin{figure} 
    \centering
    \includegraphics[width=0.6\textwidth]{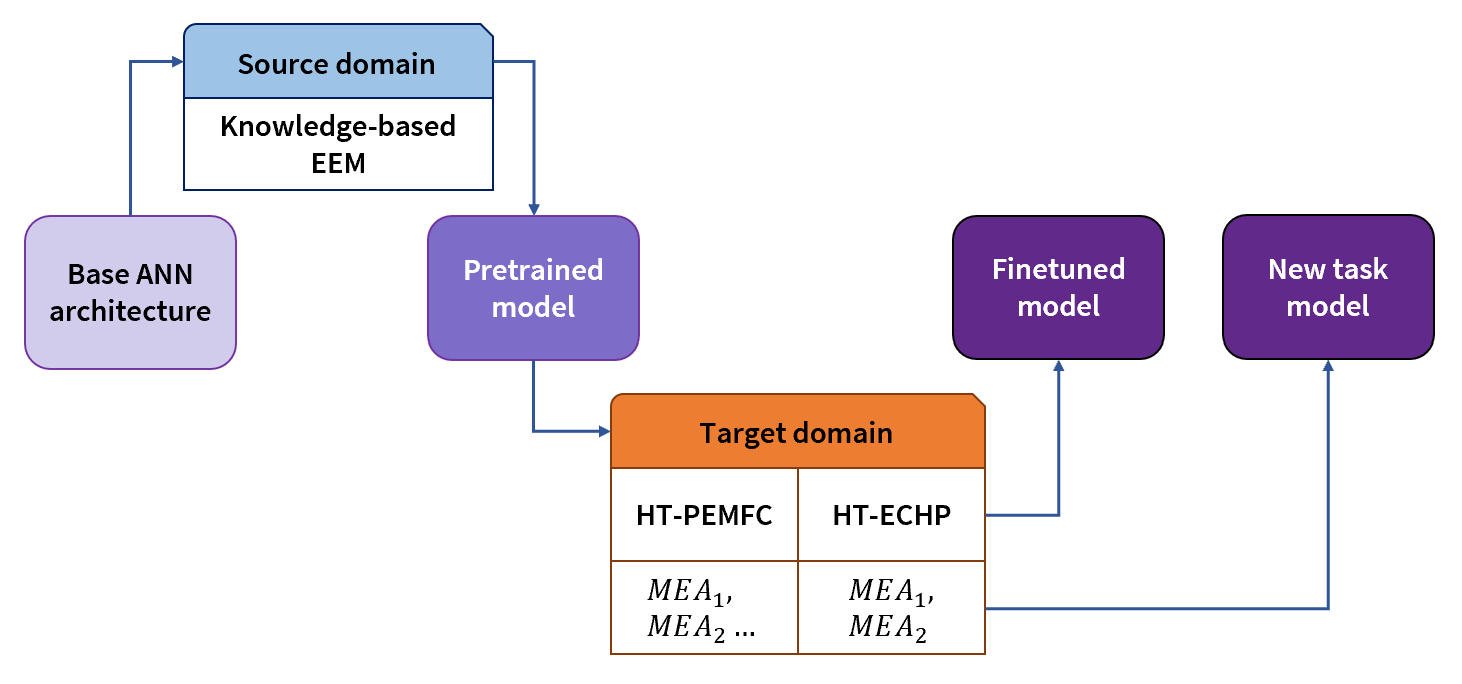}
    \caption{Transfer learning-based modeling strategy.}
    \label{fig1}
\end{figure}

The overarching strategy of this implementation is described in Figure \ref{fig1}. A knowledge-based explicit equations model (EEM) was used as the source domain. Several datasets for different membrane electrode assemblies (MEAs), temperatures, and two different devices were used as target domains for a total of 8 new models produced. Two different neural network architectures were tested as basis for the source model.

\subsection{Data preparation}
\subsubsection{HT-PEMFC simulated data}
\begin{figure} 
  \centering
  \includegraphics[width=1.0\textwidth]{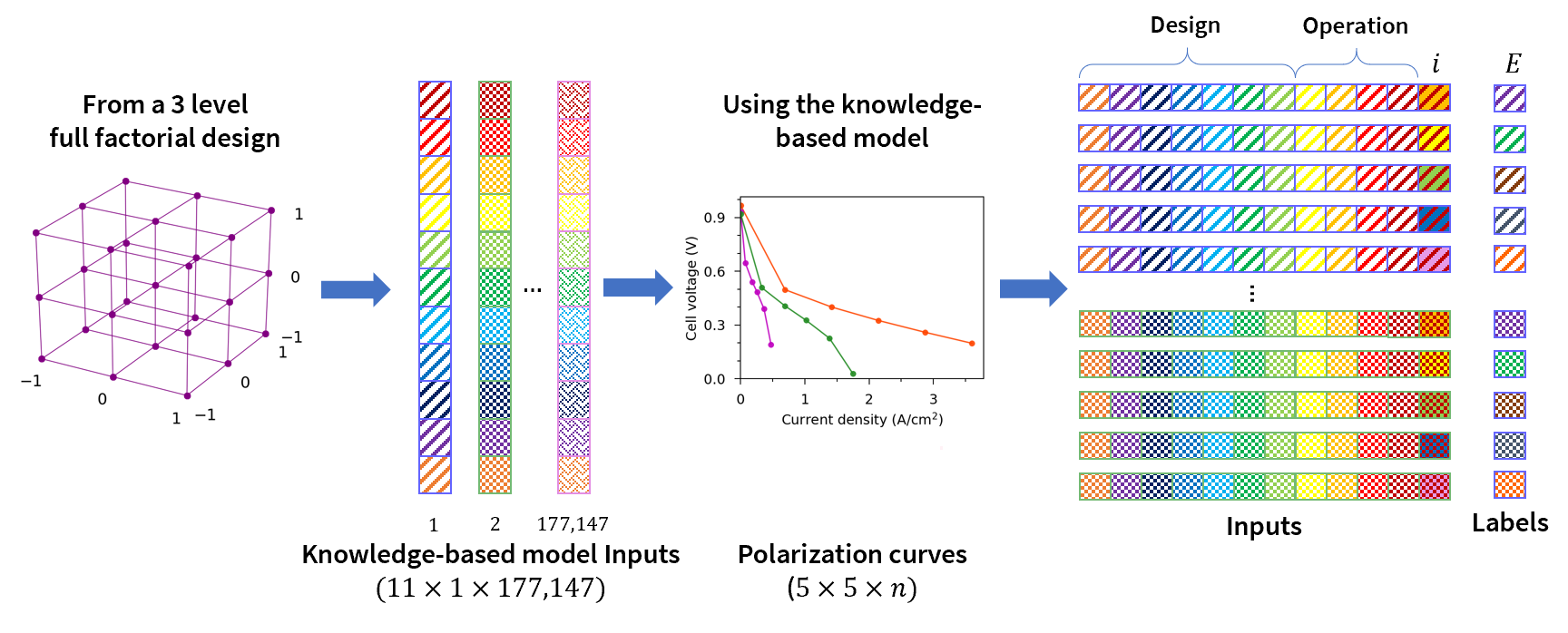}
  \caption{Simulated data generation and preprocessing strategy. The values for the inputs are structured following a \(3^{11}\) factorial design of experiments and fed into the knowledge-based model to generate 5 points in the corresponding polarization curves. The final dataset has \(177,147\times5 = 887,735\) samples with input vectors of size 12 and labels of size 1.}
  \label{fig2}
\end{figure}

For the general-training stage, simulated data was generated using the knowledge-based model reported elsewhere\cite{RN13}. Figure \ref{fig2} shows the general procedure for data preprocessing. To obtain a balanced dataset, the values for the input variables where structured following a full 3-level factorial experimental design using 11 variables (\(3^{11}\)) (see Table \ref{table1}). It is important to note that all HT-PEMFC models and data only consider pure oxygen as the oxidant. For each set of inputs, 5 points in the polarization curves were generated. The resulting datasets were arranged to generate a total of 887,735 input vectors of size 12 and the corresponding labels of size 1. Data was standardized in all cases.

\begin{table}
  \centering
  \caption{Variables and levels for the fully 3-level factorial experimental design used to structure the simulated dataset.}
  \begin{tabular}{>{\hspace{0pt}}m{0.54\linewidth}>{\hspace{0pt}}m{0.22\linewidth}>{\hspace{0pt}}m{0.15\linewidth}} 
  \toprule
  Variable                                                                        & Levels                              & Units         \\ 
  \midrule
  Hydrogen stoichiometric ratio, S\textsubscript{H\textsubscript{2}}              & [1, 1.5, 2]                         & Dimensionless \\
  Oxygen stoichiometric ratio, S\textsubscript{O\textsubscript{2}}                & [2, 2.5, 3]                         & Dimensionless \\
  Temperature, T                                                                  & [423, 463, 503]                     & K  \\
  Pressure, P                                                                     & [1, 1.5, 2]                         & atm \\
  Membrane ion exchange capacity, IEC\textsubscript{mem}                          & [1.5, 2.25, 3]                      & mequiv/g \\
  Ionomer binder ion exchange capacity, IEC\textsubscript{io}                     & [1.5, 2.25, 3]                      & mequiv/g \\
  Membrane thickness, \(\delta\)\textsubscript{mem}                               & [0.001, 0.005, 0.01]                & cm \\
  Ionomer binder thickness, \(\delta\)\textsubscript{io}                          & [\(5\times10^{7}\), 0.0001, 0.0002] & cm \\
  Carbon monoxide to hydrogen ratio in fuel stream, CO/H\textsubscript{2}         & [0, 0.05, 0.1]                      & Dimensionless \\
  Anode catalyst loading, L\textsubscript{\textsubscript{c\textsubscript{an}}}    & [0.1, 0.35, 0.6]                    & mg\textsubscript{Pt}/cm\textsuperscript{2} \\
  Cathode catalyst loading, L\textsubscript{\textsubscript{c\textsubscript{cat}}} & [0.1, 0.35, 0.6]                    & mg\textsubscript{Pt}/cm\textsuperscript{2}\\
  \bottomrule
  \end{tabular}
  \label{table1}
\end{table}

\subsubsection{Experimental data}
Table \ref{table2} summarizes the experimental data sources and its characteristics. For the HT-PEMFC models, data reported by Atanasov et al.\cite{RN5} and Venugopalan et al.\cite{RN6} were used. For the HT-PEM ECHP models, data reported by Venugopalan et al.\cite{RN7} were used. Out of the available datasets for each MEA and device, one was set aside for testing and it was not used in the training stage. Table \ref{table2} also notes the binder and HT-PEM chemistry differences. All electrocatalysts were platinum/platinum group metal (PGM) alloy nanoparticles decorated on high-surface area graphitic carbon supports.

\begin{sidewaystable}
  \small
  \centering
  \caption{Experimental datasets for few-shot learning.}
  \begin{tabular}{>{\centering\hspace{0pt}}m{0.07\linewidth} 
                  >{\hspace{0pt}}m{0.02\linewidth} 
                  >{\hspace{0pt}}m{0.02\linewidth} 
                  >{\centering\hspace{0pt}}m{0.05\linewidth} 
                  >{\centering\hspace{0pt}}m{0.03\linewidth} 
                  >{\centering\hspace{0pt}}m{0.075\linewidth} 
                  >{\centering\hspace{0pt}}m{0.05\linewidth} 
                  >{\centering\hspace{0pt}}m{0.07\linewidth} 
                  >{\centering\hspace{0pt}}m{0.05\linewidth} 
                  >{\centering\hspace{0pt}}m{0.04\linewidth} 
                  >{\centering\hspace{0pt}}m{0.04\linewidth} 
                  >{\centering\hspace{0pt}}m{0.075\linewidth} 
                  >{\centering\hspace{0pt}}m{0.07\linewidth} 
                  >{\centering\arraybackslash\hspace{0pt}}m{0.08\linewidth}} 
  \toprule
  ID &  S\textsubscript{H\textsubscript{2}} &  S\textsubscript{O\textsubscript{2}} & T    (°C) & P (atm) & HT-PEM\tablefootnote{HT-PEM chemistries - PA-PBI: Phosphoric acid (PA) imbibed polybenzimidazole (PBI), PA-QPPSf-PBI: PA imbibed quaternary benzyl pyridinium Udel® poly(arylene ether sulfone) (QPPSf)-PBI, PA-QAPOH: PA imbibed quaternary alkyl ammonium poly(phenylene), PA-TPP/Nafion: PA imbibed tin pyrophosphate-Nafion® composite} &  IEC\textsubscript{mem}\tablefootnote{EC\textsubscript{mem} is based upon the number of H\textsubscript{3}PO\textsubscript{4} mequiv per weight of H\textsubscript{3}PO\textsubscript{4} imbibed HT-PEM otherwise noted} (mequiv/g) & Electrode ionomer binder\tablefootnote{Ionomer binder chemistries - PTFSPA: poly(tetrafluorostyrene phosphonic acid-co-pentafluorostyrene), PA-QASOH: PA imbibed quaternary benzyl ammonium polystyrene, PA-QPPSf: PA imbibed QPPSf} &  IEC\textsubscript{io}\tablefootnote{IEC\textsubscript{io} is based upon H\textsubscript{3}PO\textsubscript{4} or phosphonic acid mequv per weight of polymer (including acid imbibing if applicable)} (mequiv/g) &  \(\delta\)\textsubscript{mem} (cm) &\(\delta\)\textsubscript{io} (cm) & L\textsubscript{\textsubscript{c\textsubscript{an}}}
    (mg\textsubscript{PGM}/cm\(^2\)) & L\textsubscript{\textsubscript{c\textsubscript{cat}}}
    (mg\textsubscript{PGM}/cm\(^2\)) & Electrode ionomer
    binder loading (wt\%) \\ 
  \midrule
  MEA0[6] & 1.2 & 2.2 & 160, 200, 220 & 1.59 & PA-QPPSf-PBI & 7.9 & PA-QPPSf & 8.9 & 0.005 &  0.0001 & 0.5
    37\% Pt/C & 0.5
    37\% Pt/C & 30 \\ 
  \hline\hline
  MEA1[5] & 1 & 1 & 120, 160,
    200 & 1.47 & PA-PBI & 9.1 & PTFE & n/a\tablefootnote{n/a – not applicable as the value is either zero or unknown. Value was set to zero for training. \label{note5}} & 0.005 & 0.0001 & 1.0
    Pt/C & 0.75
    Pt-alloy & n/a \textsuperscript{5} \\
  MEA2[5] & 1 & 1 & 120, 160,
    200 & 1.47 & PA-PBI & 9.1 & PTFSPA & 2.2 & 0.005 & 0.0001 & 0.5
    50\% PtRu/C & 0.6
    60\% Pt/C & 10.4 \\
  MEA3[5] & 1 & 1 & 120, 160,
    200, 240 & 1.47 & PA-QAPOH & 7.1 & PA-QASOH & 1.9 & 0.004 & 0.0001 & 0.5
    50\% PtRu/C & 0.6
    60\% Pt/C & 10.4 \\
  MEA4[5] & 1 & 1 & 120, 160,
    200, 240 & 1.47 & PA-QAPOH & 7.1 & PTFSPA & 2.2 & 0.004 & 0.0001 & 0.5
    50\% PtRu/C & 0.6
    60\% Pt/C & 10.4 \\
  MEA5[5] & 1 & 1 & 120, 160,
    200, 240 & 1.47 & PA-TPP/Nafion & n/a \textsuperscript{5} & PA-QASOH & 1.9 & 0.008 & 0.0001 & 0.5
    50\% PtRu/C & 0.6
    60\% Pt/C & 10.4 \\ 
  \hline\hline
  ECHP1 [7] & 1 & - & 160, 180, 200, 220 & 1 & PA-QPPSf-PBI & 7.9 & PA-QPPSf & 8.9 & 0.005 & 0.0001 & 0.5
    37\% Pt/C & 0.5
    37\% Pt/C & 30 \\
  ECHP2 [7] & 1 & - & 160, 180, 200, 220 & 1 & PA-QPPSf-PBI & 7.9 & PTFSPA & 2.2 & 0.005 & 0.0001 & 0.5
    37\% Pt/C & 0.5
    37\% Pt/C & 30 \\
  \bottomrule
  \end{tabular}
\label{table2}
\end{sidewaystable}

\subsection{Artificial neural network architectures}
Artificial neural networks (ANN) are a ML method in which a model is built in the form of nodes connected by edges and organized in layers (also called fully connected layers) \cite{RN21}. In each node, the outputs from the nodes in the previous layers are multiplied by its corresponding weights, which are represented by the edges, added, and then transformed by an activation function to feed the next layer of nodes. ANN are trained by calculating the predicted output for a given set of inputs, computing the error E between the result and the true value (label), and adjusting the weights accordingly. The weights between nodes \(i\) and \(j\), \( w_{ij}\), are updated using the gradient descent algorithm:
\begin{equation}
  \ w_{ij}^{updated} = w_{ij} + -lr\frac{\partial E}{\partial w_{ij}}
  \label{equation1}
\end{equation}

Here, \(lr\) is the learning rate and \(\frac{\partial E}{\partial w_{ij}}\) is the partial derivative of the error between the predicted value and the label with respect to each weight. From Eq. \ref{equation1} it follows that a greater learning rate leads to a larger the change in the weights after each iteration. An extension of the gradient descent algorithm is the so-called mini-batch gradient descent (Eq. \ref{equation2}) which reduces the variance in the estimate of the gradient by processing a subset of instances (batches) at each iteration\cite{RN22}. The mini-batch gradient descent equation is shown below: 
\begin{equation}
  \ w_{ij}^{updated} = w_{ij} + -lr\frac{1}{n}\sum_{m=n\times k}^{(k+1)\times n}\frac{\partial E}{\partial w_{ij}}
  \label{equation2}
\end{equation}

Where \(n\) is the batch size, \(k=[1,N/n]\) is the number of batches for a dataset of size N. In this work, the mini-batch gradient descent algorithm was used via the Adam optimizer\cite{RN23}. An important aspect of the use of Adam is the effect of batch size during training as it has an influence on both the convergence of the training and the generalizability of the resulting model.

A challenge in the implementation of ANN relates to selecting the correct features (inputs) to be used in the model. This is important since not all variables in a system carry the same influence over a given output, and some of them may not contribute to improve the accuracy of the model and may have an opposite, undesired effect. Convolutional neural networks (CNN)\cite{RN24} build upon the basic idea of ANN and they provide a way to extract the most relevant features for a particular problem. Due to its flexibility and robustness, CNN have been deployed successfully in many engineering scenarios \cite{RN25}. The basis of CNN is the convolution operation, in which a filter or kernel is applied in the form of a moving window sliding through the input (e.g., if the input is an array of size 1×n, the kernel could have a size \(1\times(n-3))\), thus computing an output from a portion of the input at a given time\cite{RN21, RN24}. The output of a convolutional layer in a neural network represents a collection of relationships between subsets of the input. A single kernel will extract a specific feature while applying multiple kernels extract multiple features. By stacking multiple convolutional layers, the CNN gains more expressivity because subsets can interact. After the convolutional section of the neural network, the so-called pooling operation applies a criterion to reduce the size of the array being passed to the next layer (e.g., a criterion divide the array in subsets of size \(1\times2\) and keeping only the highest value within each subset)\cite{RN26}. Pooling helps reduce the amount of information being pass through the subsequent layers lowering the overall computational costs. It also provides a way to match the size of the output from the convolutional layers to the subsequent fully connected layers.

\begin{figure} 
  \centering
  \includegraphics[width=0.7\textwidth]{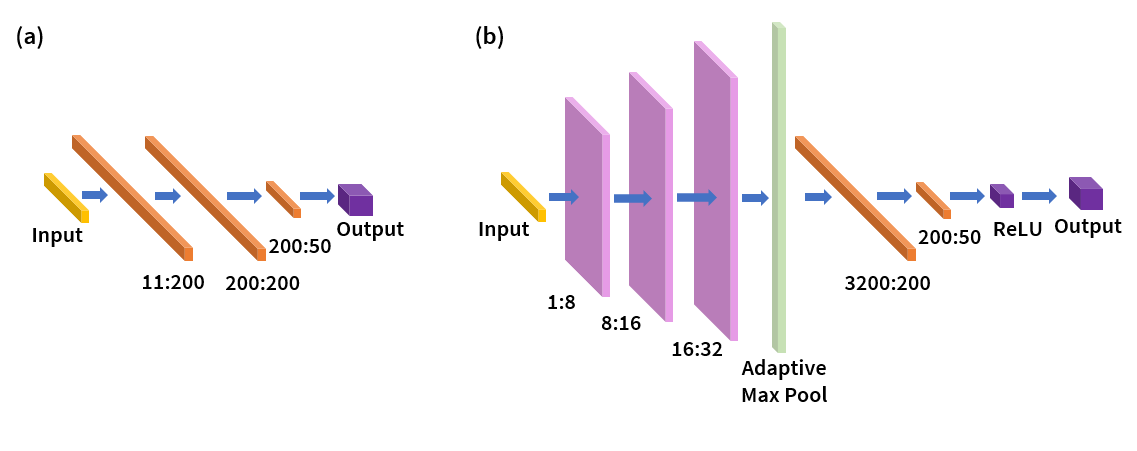}
  \caption{Schematic representation of the artificial neural network architectures. (a) FCNet: 2 fully connected hidden layers. (b) ConvNet: 3 1d convolutional layers (ReLU activation), an adaptive max pooling layer, 2 fully connected layers.}
  \label{fig3}
\end{figure}

To explore the effect of the model architecture on the accuracy of the target models, two different networks were tested: i.) a fully connected neural network (FCNet) with 2 hidden layers (11:200:50:1) and ii.) a convolutional neural network (ConvNet) with 4 convolutional layers, that features an adaptive max pooling layer and 2 fully connected layers. Figure \ref{fig3} shows an illustration of the model architectures. Both model architectures were implemented using Pytorch\cite{RN27}.

\subsection{Few-shot learning}
FSL strategies are broadly based upon the notion of preserving some of the information obtained through a first stage of training (i.e., a source model) and reusing it in some fashion to develop a target model. In this work, FSL was implemented through two different approaches: regularization and addition of layers to the source model. The description of the methods employed are provided below:

\subsubsection{Regularization}
\begin{figure} 
  \centering
  \includegraphics[width=0.9\textwidth]{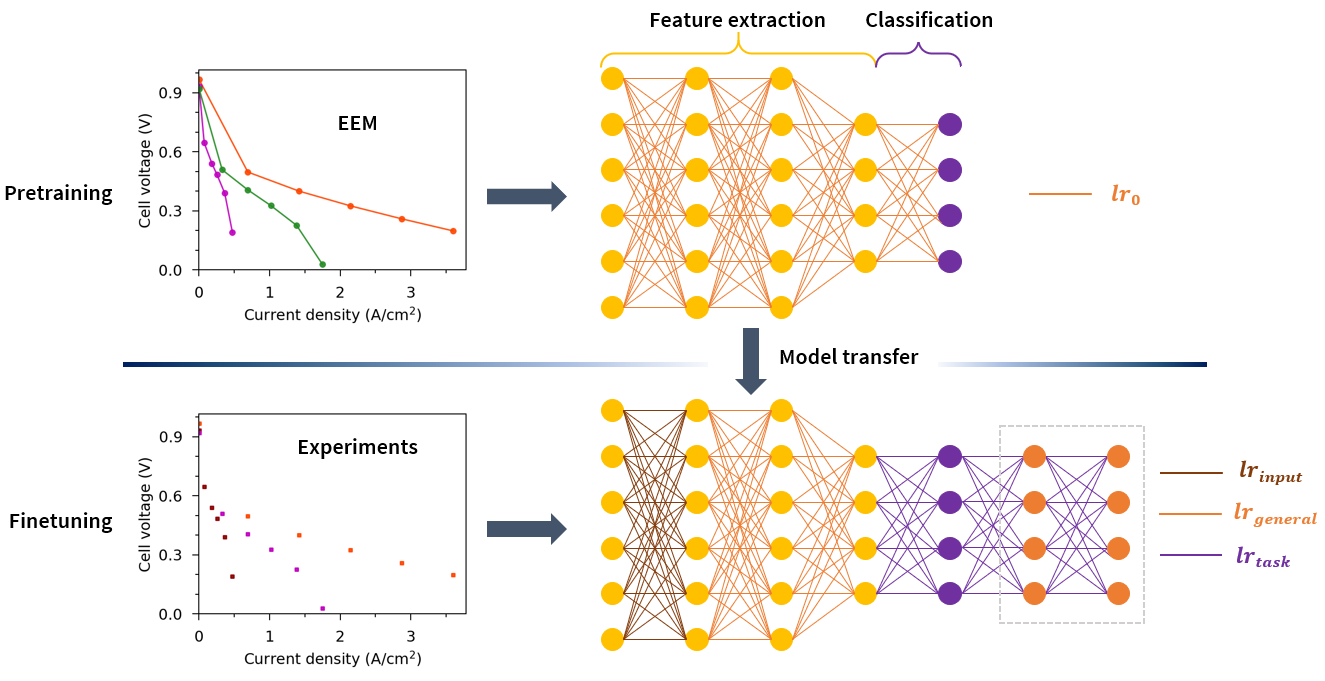}
  \caption{Schematic representation of the transfer learning approach for model development. Top row: pretraining using simulated data and a uniform learning (\(lr_0\)). Bottom row: finetuning using experimental data and a differential learning rate considering 3 sections (\([lr_{input},lr_{general},lr_{task}]\)).}
  \label{fig4}
\end{figure}

In the regularization approach, the target model has the same architecture as the source model (see Figure \ref{fig4}), and the learned parameters (weights) are adjusted during a new training stage but some restriction is applied to prevent the model from overfitting the new data\cite{RN16}. In ANN, one form of regularization involves freezing most of the weights of the model and allowing only the weights of the upper layers of the model to be updated using the new data. While this approach works in some cases, figuring out which layers to freeze can be difficult. Another form of regularization, called differential learning rate, provides a robust yet practical approach. In differential learning rate, all weights are updated during the task-specific training stage, but at a different rate (see Figure \ref{fig4}) and this helps control the influence of the new data over the model. 

\subsubsection{Addition of layers for new task learning}
For the case of learning a new task, the architecture of the source model may not be enough to represent the new domain. In such scenario, adding new layers to the source model helps attain enough expressivity to learn the new task. Figure \ref{fig4} provides an illustration of this approach. In this work, two new fully connected layers were added to the base model. Then, the differential learning rate approach was used to update the weights of the resulting model.

\section{Results and Discussion}

\begin{figure} 
  \centering
  \includegraphics[width=0.7\textwidth]{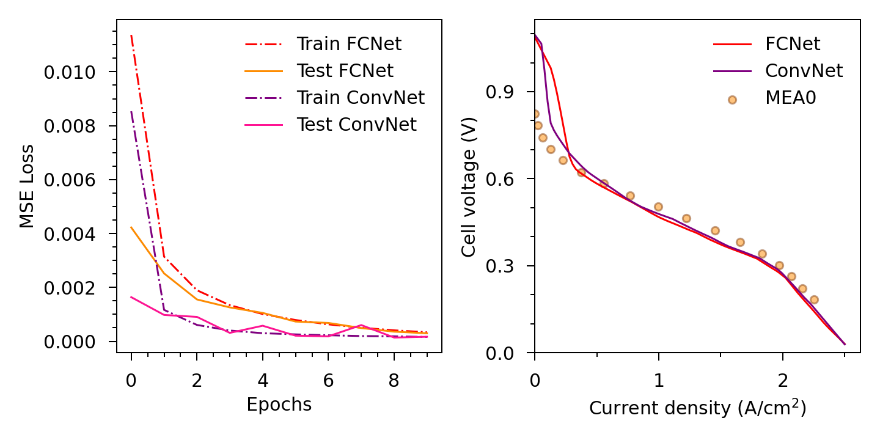}
  \caption{Training results for the source model using simulated data form a knowledge based model and two different network architectures (FCNet and ConvNet). (Left) model convergence during training and testing using the simulated dataset and (right) predictions for the MEA0 at 220 °C.}
  \label{fig5}
\end{figure}

Figure \ref{fig5} shows the training and testing loss convergence for the source model and the predictions for the MEA0 at 220 °C for the FCNet and ConvNet architectures. For both architectures, the loss function decays smoothly, and the train and testing losses reach similar values which serves as an indication that the models do not overfit or underfit the training data. As for the predictions, both architectures successfully predict the shape of the polarization curve, although the ConvNet architecture seems to capture more detail - especially at low current density values. These two models serve as source models for subsequent transfer learning activities.

In FSL for finetuning, the target domain is closely related to the source domain, as it is the case of modeling HT-PEMFCs with different MEAs and their corresponding polarization curve data sets. In several of these data sets, the materials of the MEAs are not the same (e.g., electrode binder, electrocatalyst loading and HT-PEM type); and as such, the different MEA compositions have a profound impact on single-cell polarization as well as power density. To exploit the recent availability of new HT-PEMFC data with ion-pair HT-PEMs and other known HT-PEMs, as well as phosphonated ionomer electrode binders and acid imbibed ionomer electrode binders, the source model, which was trained with data generated from the knowledge-based model, was transferred to a target model which was in turn trained with the specific new dataset. 

\begin{figure} 
  \centering
  \includegraphics[width=0.87\textwidth]{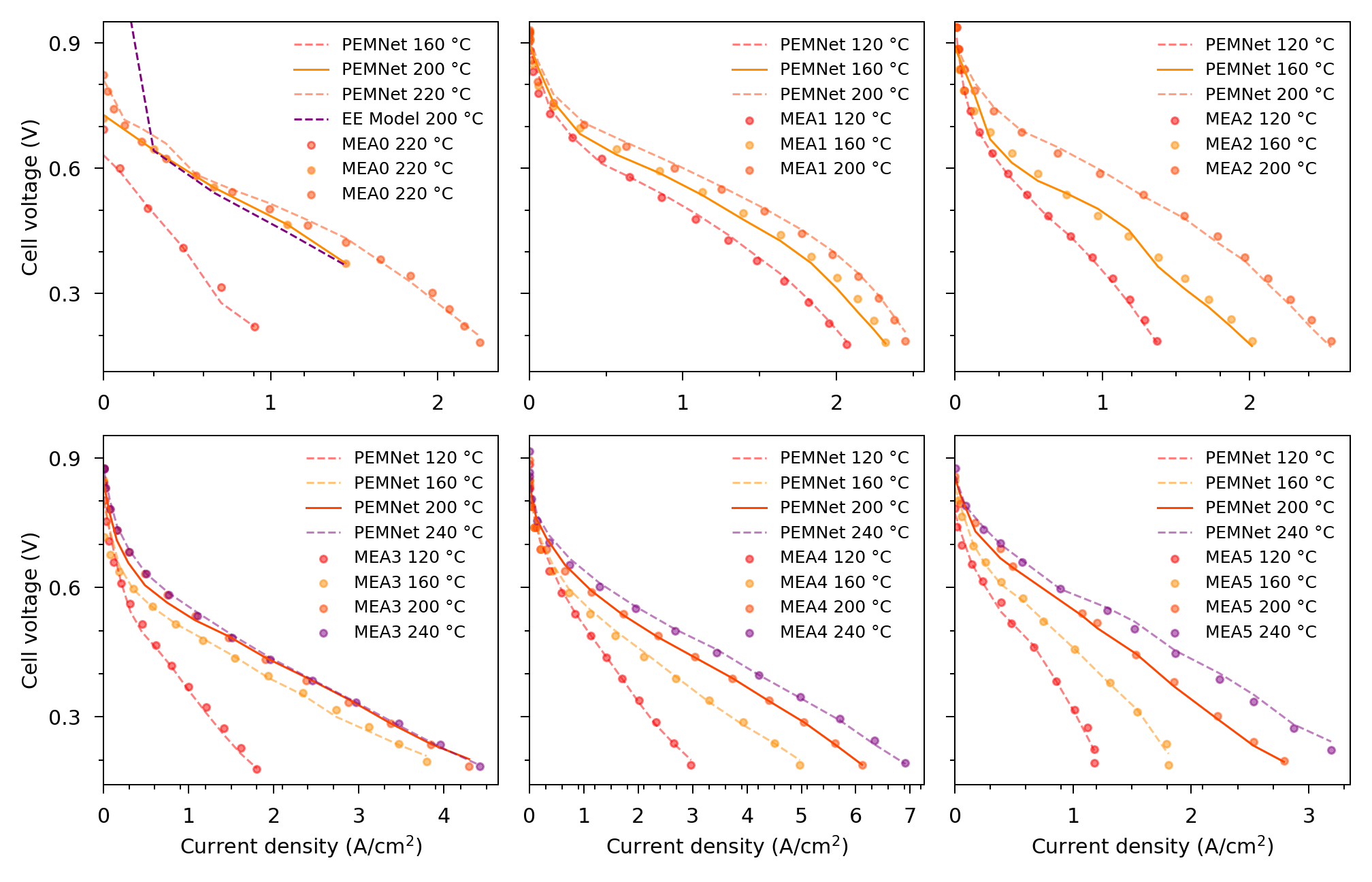}
  \caption{Model fitting for the experimental fuel cell polarization data in \cite{RN5, RN6} using {\sc PemNet} with ConvNet as source model.}
  \label{fig6}
\end{figure}

Figure \ref{fig6} shows the PEMNET predictions and the experimental data for different MEAs and different temperatures for the ConvNet source model. Table 3 shows the performance of the models obtained from the FCNet and ConvNet source models, with relative root squared mean errors (rRMSE) below 6.5\% for all cases except for MEA3 at 160 °C. The plot in the first row-first column in Figure \ref{fig6} also shows a comparison between the knowledge-based model (EEM) and {\sc PemNet}. Notably the rRMSE for the MEA0 with the ConvNet model is much lower than that of the EEM (1.04\% and 36.0\% respectively). This shows that a lower fidelity knowledge-based model can be improved using a FSL approach. This opens the opportunity for a faster development of models for new materials and device designs, helping guide new materials by identifying the optimal properties for enhancing power density in addition to other optimization activities (e.g., identifying operating parameters that maximize power and fuel efficiency constrained to PGM loading in the MEA). 

The calculated accuracy for the testing sets using the ConvNet architecture as the base model was better than that of FCNet in most cases. The details of the finetuning parameters are shown in Table \ref{table3}. The better accuracies for the ConvNet architecture could be explained by the feature extraction section provided by the convolutional layers, which are not entirely lost at the task-specific training stage. For the FCNet base model, most of the network is updated, thus losing information from the pretraining stage.

\begin{table}
\centering
\caption{Learning rates (\([lr_{input};  lr_{general};  lr_{task}]\)) and error as relative root mean squared error (rRMSE) for the few-shot learning finetuning. Testing sets are bold-faced. The best performance for each MEA is marked with (*).}
\begin{tabular}{>{\centering\hspace{0pt}}m{0.08\linewidth}>{\centering\hspace{0pt}}m{0.085\linewidth}>{\centering\hspace{0pt}}m{0.18\linewidth}>{\centering\hspace{0pt}}m{0.07\linewidth}>{\centering\hspace{0pt}}m{0.07\linewidth}>{\centering\hspace{0pt}}m{0.08\linewidth}>{\centering\hspace{0pt}}m{0.08\linewidth}>{\centering\hspace{0pt}}m{0.07\linewidth}>{\centering\arraybackslash\hspace{0pt}}m{0.065\linewidth}} 
\toprule
\multirow{2}{\linewidth}{\Centering{}Dataset} & \multirow{2}{1\linewidth}{\Centering{}Base model} & \multirow{2}{1\linewidth}{\Centering{}Differential learning rate scheme} & \multirow{2}{\linewidth}{\hspace{0pt}\Centering{}Batch size} & \multicolumn{5}{>{\Centering\hspace{0pt}}m{0.4\linewidth}}{rRMSE} \\ 
\cmidrule{5-9}
 &  &  &  & 120 °C & 160 °C & 200 °C & 220 °C & 240 °C \\ 
\midrule
MEA0 & ConvNet & {[}1e-8, 8e-6, 2e-4] & 5 & -       & 2.67\%  & \textbf{1.04\%*}  & 4.12\% & - \\
MEA0 & FCNet & {[}1e-8, 2e-4]         & 5 & -       & 2.67\% & \textbf{1.10\%}   & 4.98\% & - \\
MEA0 & EEM                            & - & -       & -       & -                 & \textbf{36.0\%} & - & - \\ 
\hline\hline
MEA1 & ConvNet & {[}1e-8, 3e-4, 5e-3] & 9 & 2.25\%  & \textbf{3.66\%*}            & 2.47\% & - & - \\
MEA1 & FCNet & {[}1e-8, 5e-3]         & 9 & 2.80\%  & \textbf{6.35\%}             & 2.93\% & - & - \\ 
\hline\hline
MEA2 & ConvNet & {[}1e-8, 8e-5, 8e-4] & 9 & 3.32\%  & \textbf{3.51\%}            & 1.47\% & - & - \\
MEA2 & FCNet & {[}1e-8, 8e-4]         & 9 & 2.16\%  & \textbf{3.40\%*}             & 1.98\% & - & - \\ 
\hline\hline
MEA3 & ConvNet & {[}1e-8, 1e-6, 8e-4] & 9 & 2.75\%  & 5.86\%                      & \textbf{3.26\%*} & -  & 1.79\% \\
MEA3 & FCNet & {[}1e-8, 8e-4]         & 9 &1.82\%  & 5.39\%                      & \textbf{5.74\%} & -   & 1.22\% \\ 
\hline\hline
MEA4 & ConvNet & {[}1e-8, 1e-6, 8e-4] & 9 & 3.46\%  & 3.01\%                      & \textbf{3.07\%*} & -  & 2.67\% \\
MEA4 & FCNet & {[}1e-8, 8e-4]         & 9 & 2.68\%    & 1.54\%                      & \textbf{4.75\%} & -   & 3.10\% \\ 
\hline\hline
MEA5 & ConvNet & {[}1e-8, 1e-6, 8e-4] & 5 & 2.94\%  & 2.80\%                      & \textbf{3.23\%*} & -  & 1.36\% \\
MEA5 & FCNet & {[}1e-8, 8e-4]         & 5 & 2.50\%  & 1.82\%                      & \textbf{3.73\%} & -    & 0.84\% \\
\bottomrule
\end{tabular}
\label{table3}
\end{table}

\begin{figure} 
    \centering
    \includegraphics[width=0.7\textwidth]{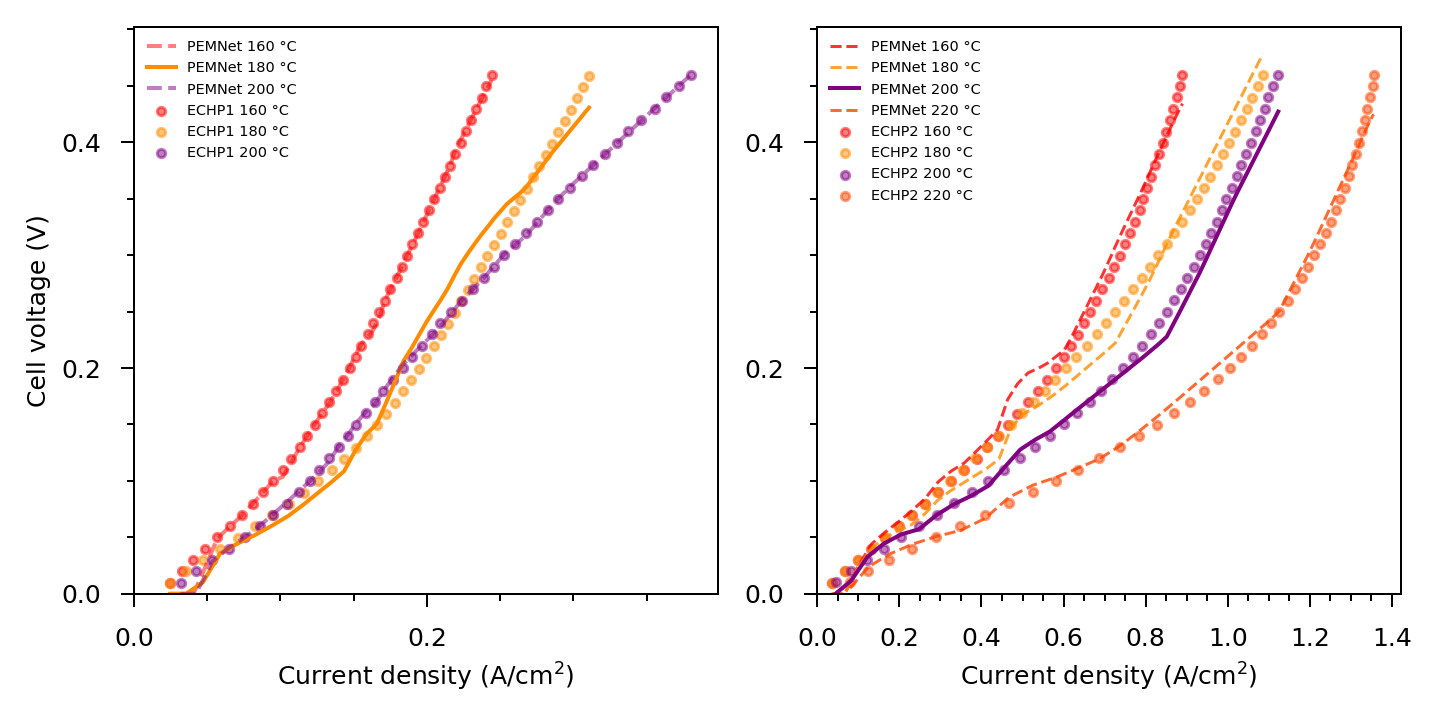}
    \caption{Model fitting for the experimental electrochemical hydrogen pump polarization data in using {\sc PemNet} with ConvNet as source model.}
    \label{fig7}
\end{figure}

Modifying the model originally developed for an HT-PEMFC to represent an HT-PEM ECHP using the same materials is pretty straightforward but nuanced differences in operating conditions (e.g., electrolytic mode as opposed to galvanic mode) as well as anode and cathode stream compositions need to be accounted for.  In this case, learning a new task strategy was pursued and it involves a more aggressive transformation of the source model\cite{RN16}. Figure \ref{fig7} shows the {\sc PemNet} predictions for the two HT-PEM ECHP devices and Table \ref{table4} shows the accuracies for each case. Although the rRMSE values for these models are slightly higher than those observed in the finetuning cases, the resulting model is still accurate and provides a good representation of the single-cell ECHP polarization behavior. For the new-task learning approach for the HT-PEM ECHP, as in the finetuning application for the HT-PEMFC, the convolutional neural network also outperformed the fully connected network.  

\begin{table}
\centering
\caption{Learning rates (\([lr_{input};  lr_{general};  lr_{task}]\)) and error as relative root mean squared error (rRMSE) for the few-shot learning of a new task. Testing sets are bold-faced. The best performance for each MEA is marked with (*).}
\begin{tabular}{>{\centering\hspace{0pt}}m{0.08\linewidth}>{\centering\hspace{0pt}}m{0.085\linewidth}>{\centering\hspace{0pt}}m{0.18\linewidth}>{\centering\hspace{0pt}}m{0.1\linewidth}>{\centering\hspace{0pt}}m{0.086\linewidth}>{\centering\hspace{0pt}}m{0.086\linewidth}>{\centering\hspace{0pt}}m{0.086\linewidth}>{\centering\arraybackslash\hspace{0pt}}m{0.08\linewidth}} 
\toprule
\multirow{2}{1\linewidth}{\Centering{}Dataset} & \multirow{2}{1\linewidth}{\Centering{}Base model} & \multirow{2}{1\linewidth}{\Centering{}Differential learning rate scheme} & 
\multirow{2}{1\linewidth}{\Centering{}Batch size} & \multicolumn{4}{>{\Centering\hspace{0pt}}m{0.338\linewidth}}{rRMSE} \\ 
\cmidrule{5-8}
 &  &  &  & 160 °C & 180 °C & 200 °C & 220 °C \\ 
\midrule
ECHP1 & ConvNet & {[}4e-4, 1e-2, 6e-2]  & 80  & 2.69\% & \textbf{ 8.46\%* } & 1.48\% & - \\
ECHP1 & FCNet & {[}4e-4, 6e-2]          & 80  & 1.20\%  & \textbf{ 11.03\% } & 0.95\% & - \\ 
\hline\hline
ECHP2 & ConvNet & {[}4e-4, 1e-2, 6e-2]  & 80  & 5.55\%  & 6.50\% & \textbf{ 6.38\%* } & 4.61\% \\
ECHP2 & FCNet & {[}4e-4, 6e-2]          & 80 & 2.08\%  & 2.25\% & \textbf{ 19.15\% } & 2.83\% \\
\bottomrule
\end{tabular}
\label{table4}
\end{table}

\begin{figure} 
  \centering
  \includegraphics[width=0.5\textwidth]{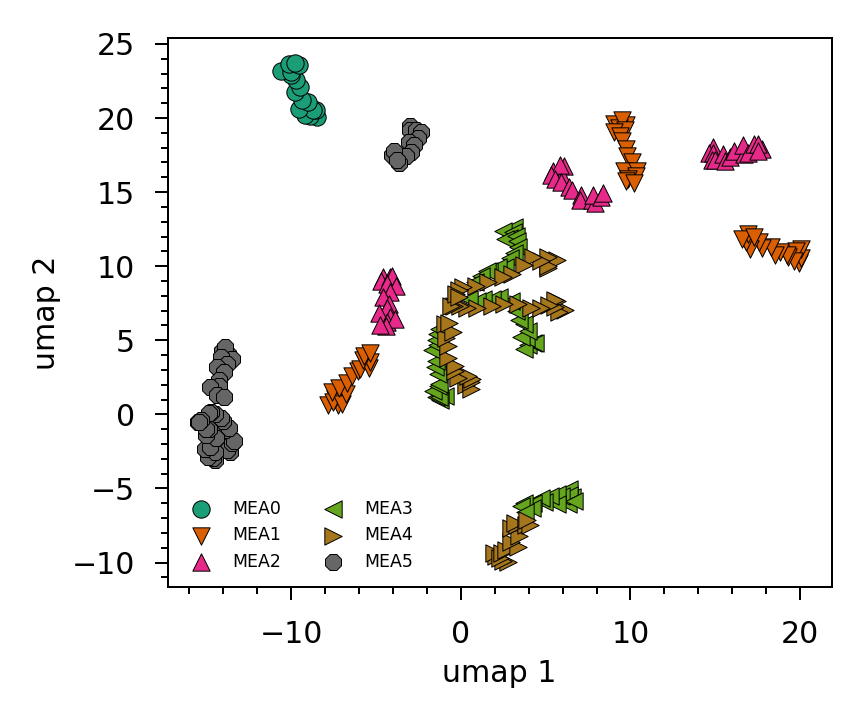}
  \caption{Two-dimensional projection of all the datasets using UMAP.}
  \label{fig8}
\end{figure}

An important question in TL relates to the difficulty for the method to accurately learn the new model. To explore this idea, a 2-dimensional projection of the original datasets provides a useful visualization of the differences between them. Figure 8 shows the projection of the experimental datasets via Uniform Manifold Approximation and Projection (UMAP)\cite{RN27} using cosine as distance metric so that the distances observed in the project can be correlated to the actual distances in the high-dimensional space. In Figure  \ref{fig8} the fuel cell data is projected as compact clusters while the ECHP data is projected as highly disperse clusters. At the same time, the ECHP models required a much higher batch size, in addition to the expected higher learning rate. These observations suggest batch size plays a very important role in FSL, with higher values being necessary because the differences between the source and target models, as well as the sparsity of the data. These latter observations agree with what has been reported previously\cite{RN21} but for different engineering applications.

\section{Conclusions}
{\sc PemNet}, a Transfer Learning-based approach for modeling high-temperature electrochemical systems was introduced and its applicability was explored for two different single-cell devices, HT-PEMFCs and HT-PEM ECHPs, with different polymer electrolyte MEA chemistries and various cell operating conditions. The implementation uses a knowledge-model to generate data and pretrain a source model. To obtain a target model, a differential learning rate FSL implementation was used. This enables the faster development of models and has implications for the proliferation of HT-PEM fuel cells and electrochemical hydrogen pumps in the global market. Furthermore, for the cases studied here, using convolutional neural networks provides an advantage over fully connected networks. Finally, it was also found that a notion of distance between the different systems, obtained from the low dimensional projection of the dataset, can inform the necessary batch size to be used for mini-batch gradient descend.

\section{Aknowledgements}
This material is based upon work supported by the U.S. Department of Energy’s Office of Energy Efficiency and Renewable Energy (EERE) under the Advanced Manufacturing Office (AMO) Award Number DE-EE0009101. This report was prepared as an account of work sponsored by an agency of the United States Government. Neither the United States Government nor any agency thereof, nor any of their employees, makes any warranty, express or implied, or assumes any legal liability or responsibility for the accuracy, completeness, or usefulness of any information, apparatus, product, or process disclosed, or represents that its use would not infringe privately owned rights. Reference herein to any specific commercial product, process, or service by trade name, trademark, manufacturer, or otherwise does not necessarily constitute or imply its endorsement, recommendation, or favoring by the United States Government or any agency thereof. The views and opinions of authors expressed herein do not necessarily state or reflect those of the United States Government or any agency thereof. Luis A. Briceno-Mena thanks the support received from Universidad de Costa Rica.

\bibliographystyle{unsrt}  
\bibliography{references} 

\end{document}